\documentclass[10pt,twocolumn,letterpaper]{article}

\usepackage[pagenumbers]{wacv} 

\usepackage{graphicx}
\usepackage{amsmath}
\usepackage{amssymb}
\usepackage{booktabs}
\usepackage{times}
\usepackage{epsfig}
\usepackage{array}

\usepackage[pagebackref,breaklinks,colorlinks]{hyperref}

\usepackage[capitalize]{cleveref}
\crefname{section}{Sec.}{Secs.}
\Crefname{section}{Section}{Sections}
\Crefname{table}{Table}{Tables}
\crefname{table}{Tab.}{Tabs.}

\begin{document}

\title{GTA: Guided Transfer of Spatial Attention from Object-Centric Representations}

\author{%
  SeokHyun Seo$^{1}$\thanks{Equal contribution}\ \ \ 
  Jinwoo Hong$^{1}$\footnotemark[1] \ \ \
  JungWoo Chae$^{1}$\footnotemark[1] \ \ \
  Kyungyul Kim$^{1}$\ \ \
  Sangheum Hwang$^{1,2}$\thanks{Corresponding author}\\
  \\
    $^{1}$LG CNS AI Research, Seoul, South Korea\\
    $^{2}$Seoul National University of Science and Technology, Seoul, South Korea\\
  \tt\small\texttt{\{serereuk186, cjwoolgcns, jinwoo.hong, kyungyul.kim, shwang\}@lgcns.com}
}
\maketitle

\begin{abstract} 
   Utilizing well-trained representations in transfer learning often results in superior performance and faster convergence compared to training from scratch. However, even if such good representations are transferred, a model can easily overfit the limited training dataset and lose the valuable properties of the transferred representations. This phenomenon is more severe in ViT due to its low inductive bias. Through experimental analysis using attention maps in ViT, we observe that the rich representations deteriorate when trained on a small dataset. Motivated by this finding, we propose a novel and simple regularization method for ViT called Guided Transfer of spatial Attention (GTA). Our proposed method regularizes the self-attention maps between the source and target models. A target model can fully exploit the knowledge related to object localization properties through this explicit regularization. Our experimental results show that the proposed GTA consistently improves the accuracy across five benchmark datasets especially when the number of training data is small. 
\end{abstract}

\section{Introduction}

\label{sec:intro}
The Vision Transformer (ViT) has demonstrated impressive performance in a variety of computer vision tasks such as image classification~\cite{vit, cait, deit, deit3, swin, tnt, swin2}, segmentation~\cite{deit3, swin, swin2, tnt}, object detection~\cite{swin, swin2, tnt}, and image generation~\cite{gpt, parmar2018image, styleswin2}, surpassing traditional convolutional neural networks (CNNs). Unlike CNNs that rely entirely on convolution operations which are designed to capture locality, neighborhood structure, and translation equivariance, only the multi-layer perceptron (MLP) component in ViT is responsible for learning those characteristics. The main difference between ViT and CNNs is the self-attention mechanism in the multi-head self-attention (MSA) layer, which globally aggregates spatial features from input tokens with normalized importance~\cite{vit}. ViT is known to have a lower inductive bias compared to CNNs, meaning that it requires more training data to obtain a well-performing model. As a result, when the available training data is limited, ViT generally shows lower performance than CNNs~\cite{lee2021vision}. In a recent study~\cite{howdo}, the authors argued that MSA has both advantages and disadvantages. The advantage is its ability to flatten the loss landscape, which can improve accuracy and robustness in large data regimes. On the other hand, the disadvantage is that MSA allows the negative Hessian eigenvalues when trained on limited training data. These negative Hessian eigenvalues can lead to a non-convex loss landscape, which can disturb model training. The study also demonstrated that self-attention can be interpreted as a \textit{large-sized} and \textit{data-specific} spatial kernel \cite{howdo}. 

\begin{figure}[t!]
    \centering
    \includegraphics[width=8.3cm]{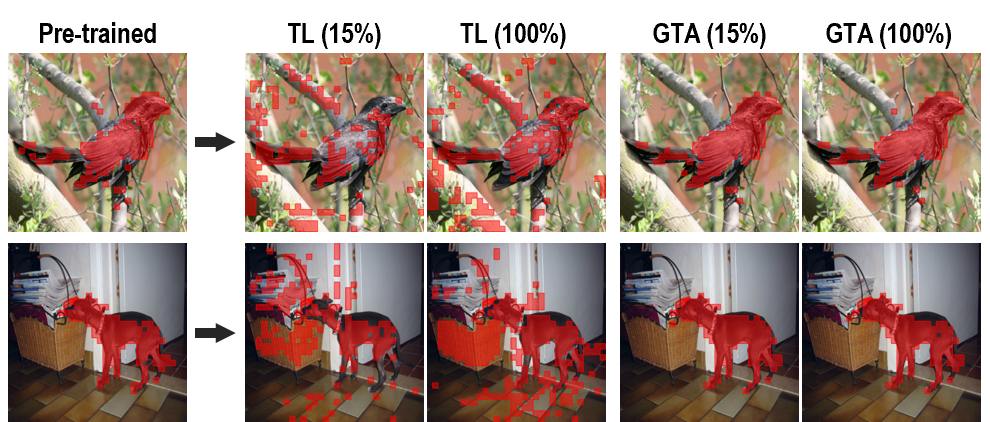}
   \caption{\textbf{Comparison of self-attention maps from pre-trained, na\"{\i}vely fine-tuned, and GTA-traind models.} The self-attention maps of the multiple heads are aggregated with max values, and visualized in red color. Each column shows the attention maps from the models that are pre-trained, fine-tuned, and fine-tuned with GTA on 15\% and 100\% of training data, respectively. GTA shows that it is capable of fully leveraging well-trained representations learned by the upstream task.}
   \vspace{-10pt}
\label{fig1_intro}
\end{figure}

When training data is scarce, transfer learning (TL) has been considered as the de-facto paradigm in practice. Pre-trained models, which have been trained with large-scale datasets, have enabled faster training and high generalization performance in TL scenarios. Various TL techniques have been proposed to effectively learn target tasks by utilizing well-trained representations transferred from pre-trained models~\cite{l2, l2sp, bss, co-tuning, 3things}. Recently, self-supervised learning (SSL) has emerged as a promising approach for learning visual representations without using class labels. SSL allows to obtain domain-specific representations by training an unlabeled large-scale dataset related to the target domain of interest, e.g., SSL on large-scale medical images~\cite{azizi2021big}. With this advantage, SSL can serve as a powerful alternative to supervised learning (SL) to address the domain discrepancies in various TL scenarios. The ViT architecture has recently proven advantageous for SSL due to its ability to fully leverage large-scale datasets. In particular, some studies have shown high TL performance by utilizing accurate object-centric representation features, which can also be helpful for semantic segmentation~\cite{dino,ibot}. 

When applying commonly used TL techniques to ViT, the object-centric representations from well-trained models may deteriorate. We experimentally confirmed that the quality of well-trained features deteriorates after fine-tuning based on the visualization of self-attention maps from na\"{\i}vely fine-tuned ViT models, and assessed the influence of the amount of training data (see Figure~\ref{fig1_intro}). Through the self-attention maps, we can visually see which image tokens are particularly attended to perform the target task. As shown in Figure~\ref{fig1_intro}, the visualization results indicate that ViT trained with basic fine-tuning tends to learn shortcuts, e.g., the features corresponding to the background (i.e., non-object area). Such shortcut learning is an undesirable behavior due to the correlation between objects and background in few-shot settings, which hinders generalization~\cite{rectify, rectifying}. Even with a relatively sufficient amount of training data, ViT still focuses on non-object regions due to its low inductive bias. Motivated by this observation, we hypothesize that TL performance can be improved if we can prevent the degradation of attention quality of pre-trained SSL models.

In this paper, to address this issue, we propose the Guided Transfer of spatial Attention (GTA) method, which effectively leverages pre-trained knowledge containing discriminative attention to enhance the TL performance of ViT, even with the limited size of the training dataset. Specifically, we explicitly regularize the self-attention logits of a downstream network (i.e., a target network) through a simple squared $L_2$ distance. Using various benchmark datasets, we compare our proposed GTA with existing TL methods including a method specifically designed for ViT~\cite{3things} to demonstrate its superiority over comparison targets. 
To evaluate the effectiveness and importance of guiding self-attention, we compare the performance of guiding other output features from ViT, e.g., outputs of MSA layers or transformer blocks. In addition, we experimentally evaluate whether we can expect a performance boost when GTA is used in conjunction with TransMix~\cite{transmix}, a label-mixing augmentation method specifically designed for ViT based on attention scores. It differs from Mixup~\cite{mixup} and CutMix~\cite{cutmix} which determine augmented labels based on randomly sampled mixing coefficients between two images. Finally, we evaluate the factors that may affect the performance of GTA including the use of SL as a guide model.

Our main contribution can be summarized as follows:
\begin{itemize}
  \item We propose a simple yet effective TL technique for ViT named GTA. Our proposed GTA effectively improves performance by explicitly guiding one of the MSA components, self-attention logits. 
  \item We demonstrate that as the amount of training data decreases, the likelihood of self-attention deviating from the pre-trained model and concentrating on non-object regions increases. Our experimental results show the critical importance of guiding self-attention during ViT training in TL settings, especially when the amount of training data is limited.
\end{itemize}
\section{Related Work}
\begin{figure*}[h!]
    \centering
    \includegraphics[width=14.0cm]{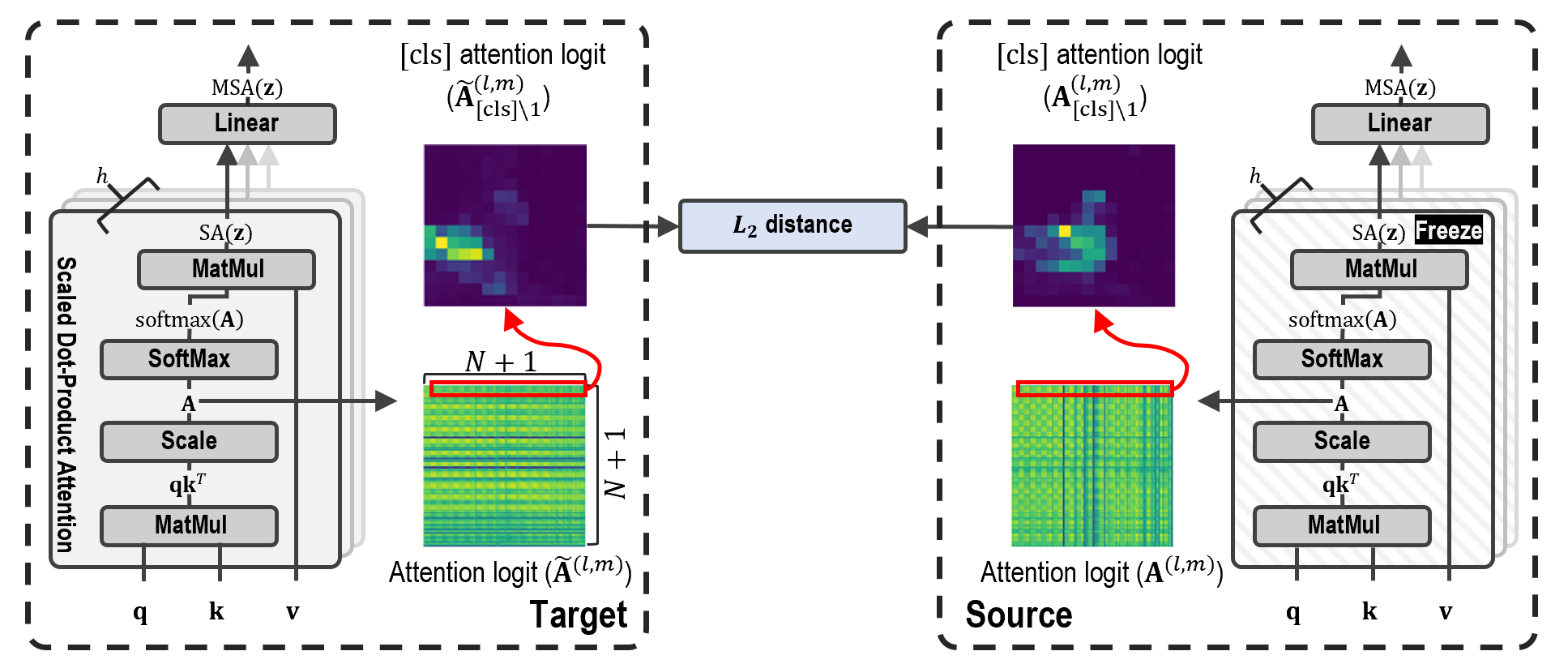}
   \caption{\textbf{The overall pipeline of the proposed GTA.} An image is first fed into both the frozen source model and the trainable target model. By minimizing the $L_2$ distance between the attention logits from each model, the target model is optimized for the current task while focusing on the image tokens that require attention by exploiting the source model.}
   \vspace{-10pt}
\label{fig_method}
\end{figure*}

\paragraph{Transfer learning.}
TL is the most common and popular method in deep learning that can be applied to various downstream tasks~\cite{tl1, tl2}. It not only improves performance but also ensures fast convergence of training by utilizing pre-trained models~\cite{tl3}. Some studies have proposed methods to exploit the pre-trained knowledge and improve performance by regularizing features~\cite{delta, bss}. DELTA measures the importance of feature channels in the CNN model and regularizes the channels far from the pre-trained activations to leverage the transferred knowledge~\cite{delta}. BSS shows that small eigenvalues of transfer features cause negative transfer, and penalizing small eigenvalues during TL to suppress untransferable spectral components can improve performance~\cite{bss}. Another method of exploiting prior knowledge is weight-based regularization, which controls the weight changes during downstream training~\cite{l2,l2sp}. $L_2$ regularization penalizes changes in model weights~\cite{l2}, and $L_2$-SP utilizes $L_2$ constraints on the weights by using the pre-trained model as the starting point to leverage the learned inductive bias~\cite{l2sp}. Co-tuning~\cite{co-tuning} has shown impressive performance improvements by exploiting the label relationship between the upstream and downstream tasks. However, in this work, to ensure ease of implementation and scalability, we only focus on methods that do not require additional data~\cite{co-tuning} or pre-processing steps for training~\cite{delta}. While many studies on TL have focused on CNNs, it is shown that fine-tuning only the MSA layers can improve performance compared to full fine-tuning~\cite{3things}.

\paragraph{Self-supervised learning.}
SSL has received considerable attention due to its ability to learn meaningful representations without requiring human annotations~\cite{moco, simclr, mocov3, byol, dino, ibot, mugs, mae, msn, ssl_transfer}. This is accomplished by engaging in self-imposed pretext tasks such as contrastive learning~\cite{simclr, moco}, utilizing the teacher-student framework~\cite{dino, byol}, predicting pixels of masked patches~\cite{mae} and a combination of pretext tasks~\cite{ibot, mugs, msn}. In particular, iBOT~\cite{ibot}, shows a significant improvement in the attention quality of ViT. We focus on models pretrained using SSL due to their aforementioned advantages and popularity, but also show that our method is effective on SL models.

\paragraph{Knowledge-distillation}
Knowledge distillation (KD) is a method where a larger teacher model guides a smaller student model to achieve a similar objective of the teacher \cite{kd}. KD can be broadly categorized into logit-based and feature-based approaches. KD and transfer learning (TL) share common ground in leveraging a pre-trained model on large-scale datasets. However, while KD focuses on transferring knowledge from the teacher model to the student model, TL seeks the most effective way to exploit the knowledge of a pre-trained source model for a new target task. In this context, we introduce GTA as a novel TL methodology for ViT and compare its performance with existing TL methods.

\section{Method}
This section presents our proposed approach, which aims to fully exploit the SSL representations from ViT for effective TL to unseen target datasets. We first provide a brief summary of the computations involved in ViT and then introduce the proposed GTA method.

\subsection{Preliminaries}
ViT consists of a stack of transformer blocks, each of which contains MSA and feed-forward layers.
Let $\mathbf{z} \in \mathbb{R}^{(N+1) \times D}$ be input features of a specific transformer block, where $N$ denotes the number of input features corresponding to image patches and $D$ represents the dimensionality of features. Note that $\mathbf{z}$ has one extra dimension since the extra learnable \texttt{[cls]} token is typically used to aggregate patch-level features. The value of $N$ can be calculated as $N = HW/P^2$, where $H$ and $W$ denote the height and width of an image, respectively, and $P$ represents the size of patches. 

The MSA layer computes a weighted sum of value embeddings, where the weights are computed with query and key embeddings. For a single attention head, these embeddings are obtained by the associated weights $\mathbf{W_q}$, $\mathbf{W_k}$, and $\mathbf{W_v}$, respectively. Specifically, a query $\mathbf{q}$, a key $\mathbf{k}$, and a value $\mathbf{v}$ are given by:
\begin{align}  
    &\mathbf{q} = \mathbf{z}\mathbf{W_q},  \mathbf{k}=\mathbf{z}\mathbf{W_k}, 
    \mathbf{v} = \mathbf{z}\mathbf{W_v},
\end{align}
i.e., $\mathbf{q}$, $\mathbf{k}$, and $\mathbf{v}$ are all $(N+1)\times k$ dimensional matrices where $k$ denotes an embedding dimension of a single attention head. Typically, $k$ is set to $D/h$ when MSA has $h$ attention heads. By computing a scaled dot product between $q$ and $k$, we can obtain \textbf{the attention logit matrix $\mathbf{A}$} as follows:
\begin{align}
    &\operatorname{\mathbf{A}} = \mathbf{q}\mathbf{k}^T / \sqrt{k}, \ \ \ \operatorname{\mathbf{A}} \in \mathbb{R}^{(N+1) \times (N+1)}.
\label{eq:attention-logit}
\end{align}
It should be noted that this attention logit plays a crucial role in our GTA. Then, the output features $\operatorname{SA}(\mathbf{z}) \in \mathbb{R}^{(N+1) \times k}$ can be obtained by $\operatorname{softmax}(\operatorname{\mathbf{A}})\mathbf{v}$
where $\operatorname{softmax}(\cdot)$ applies the softmax operation to every row of a matrix.
Finally, MSA aggregates the outputs from $h$ attention heads using the weight $\mathbf{W_{proj}} \in \mathbb{R}^{(h \cdot k) \times D}$ to compute the final MSA output:
\begin{align}  
    &\operatorname{MSA}(\mathbf{z}) = [\operatorname{SA}_1(\mathbf{z}), \cdots, \operatorname{SA}_h(\mathbf{z})]\mathbf{W_{proj}}.
\end{align}
Finally, position-wise feed-forward layers are employed to generate output features $\mathbf{z}^{\prime}$ of a transformer block from $\operatorname{MSA}(\mathbf{z})$. Note that we have excluded layer normalization to simplify the explanation.

\subsection{Spatial Attention Guidance}
Inspired by the findings that ViT models pre-trained on large-scale datasets using SSL show remarkable foreground localization capabilities, and that MSA facilitates spatial mixing of input features, we propose a simple yet effective TL strategy that is tailor-made for ViT.

Given the attention logit matrix $\mathbf{A}^{(l,m)}$ (Eq.~\ref{eq:attention-logit}) of the $l$-th head in $m$-th transformer block, we focus on the attention logit values that relate to the \texttt{[cls]} token query. More specifically, given $\mathbf{A}^{(l,m)} = [\mathbf{A}^{(l,m)}_{\texttt{[cls]}}; \mathbf{A}^{(l,m)}_{1}; \cdots ; \mathbf{A}^{(l,m)}_{N}]$, we only consider the \texttt{[cls]} attention vector, excluding the first element (which is simply a scaled norm of the \texttt{[cls]} query vector), denoted as $\mathbf{A}^{(l,m)}_{\texttt{[cls]} \setminus 1}$. This attention vector contains valuable information on which input patches should be attended to perform a given task. 

Assuming that $\mathbf{A}^{(l,m)}_{\texttt{[cls]} \setminus 1}$ offers robust spatial mixing coefficients, leveraging this knowledge for TL on downstream tasks can be achieved through a straightforward implementation of constrained optimization, with the constraint that fine-tuned attention logits should be similar to those of initial models (e.g., pre-trained SSL models):
\begin{equation}
\min ~ \mathcal{L}_{\operatorname{CE}} \quad \textrm{s.t.} ~~ \mathbf{A}^{(l,m)}_{\texttt{[cls]} \setminus 1} \approx \tilde{\mathbf{A}}^{(l,m)}_{\texttt{[cls]} \setminus 1} \quad \forall ~ l,m
\end{equation}
where $\mathcal{L}_{\operatorname{CE}}$ represents the cross entropy loss and $\tilde{\mathbf{A}}$ denotes an attention logit matrix of a target model trained during fine-tuning. To this end, we employ a simple squared $L_2$ distance for the constraint. Therefore, given a coefficient $\lambda$, our objective function $\mathcal{L}$ during fine-tuning reduces to:
\begin{align} 
    &\mathcal{L} = \mathcal{L}_{\operatorname{CE}} +  
    \lambda \sum_{l,m} {\left\Vert \mathbf{A}^{(l,m)}_{\texttt{[cls]} \setminus 1} - \tilde{\mathbf{A}}^{(l,m)}_{\texttt{[cls]} \setminus 1} \right\Vert}_2^2
    \label{eqn:guide}
\end{align}

Our regularization term, GTA, can be interpreted as transferring spatial kernels from a pre-trained model to a target model. That is, the target model tries to learn how to mix channel information while preserving the similarity of spatial mixing coefficients to those of the pre-trained model. It is worth noting that although GTA is motivated by the localization property of SSL models, it is also effective in TL with SL models since it allows the target model to selectively utilize pre-trained features. 

\begin{table}[h!]
\centering
\small
\newcolumntype{L}[1]{>{\raggedright\let\newline\\\arraybackslash\hspace{0pt}}m{#1}}
\newcolumntype{C}[1]{>{\centering\let\newline\\\arraybackslash\hspace{0pt}}m{#1}}
\newcolumntype{R}[1]{>{\raggedleft\let\newline\\\arraybackslash\hspace{0pt}}m{#1}}
\begin{tabular}{L{2.8cm}R{1.4cm}R{1.2cm}R{1.2cm}}
\toprule
\textbf{Dataset}          & \textbf{\# category} & \textbf{\# train} & \textbf{\# test}  \\ \hline
CUB \cite{cub200}    & 200      & 5994        & 5794        \\
Cars \cite{stanfordcars}   & 196      & 8144        & 8041        \\
Aircraft \cite{aircraft}   & 100      & 6667        & 3333        \\
Dogs \cite{stanforddogs}   & 120      & 12000       & 8580        \\
Pet \cite{oxfordiiitpets} & 37       & 3680        & 3669        \\
\bottomrule
\end{tabular}
\vspace{-4pt}
\caption{\textbf{Overview of dataset statistics.} Table shows the number of classes, and training and test images of each dataset used in our experiments.}
\vspace{0pt}
\label{tbl:dataset}
\end{table}

\begin{table*}[!]
\begin{center}
     \small
     \newcolumntype{L}[1]{>{\raggedright\let\newline\\\arraybackslash\hspace{0pt}}m{#1}}
     \newcolumntype{C}[1]{>{\centering\let\newline\\\arraybackslash\hspace{0pt}}m{#1}}
     \newcolumntype{R}[1]{>{\raggedleft\let\newline\\\arraybackslash\hspace{0pt}}m{#1}}
     \begin{tabular}{L{1.5cm}L{4.2cm}R{2.3cm}R{2.3cm}R{2.3cm}R{2.3cm}}
         \toprule
                                 &                        & \multicolumn{4}{c}{\textbf{Sampling Rates [Acc@1]}} \\
             {\textbf{Dataset}}  & {\textbf{Method}}      & \textbf{15\%} & \textbf{30\%} & \textbf{50\%} & \textbf{100\%} \\
             \hline
             CUB            & Fine-tune (baseline)         & 41.376 ± 0.415 & 62.697 ± 0.552 & 75.158 ± 0.369 & 84.444 ± 0.166 \\ 
             ~              & $L_2$-SP \cite{l2sp}            & 41.554 ± 1.020 & 63.261 ± 0.640 & 75.371 ± 0.345 & 84.898 ± 0.274 \\ 
             ~              & BSS   \cite{bss}             & 41.382 ± 0.787 & 62.870 ± 0.343 & 75.406 ± 0.147 & 84.501 ± 0.320 \\ 
             ~              & Attention only (freeze FFN) \cite{3things} & 42.636 ± 0.582 & 62.686 ± 0.511 & 75.175 ± 0.036 & 85.048 ± 0.232 \\ 
             ~              & FFN only (freeze attention) \cite{3things} & 37.349 ± 0.901 & 58.181 ± 0.121 & 71.839 ± 0.217 & 82.902 ± 0.138 \\ 
             ~              & GTA                        & \textbf{51.525 ± 0.449} & \textbf{68.416 ± 0.419} & \textbf{78.058 ± 0.089} & \textbf{85.543 ± 0.320} \\
             \hline
             Cars           & Fine-tune (baseline)        & 56.100 ± 0.675 & 78.502 ± 0.167 & 87.091 ± 0.132 & 93.065 ± 0.093 \\ 
             ~              & $L_2$-SP    \cite{l2sp}           & 56.676 ± 0.783 & 78.713 ± 0.316 & 87.257 ± 0.168 & \textbf{93.276 ± 0.038} \\ 
             ~              & BSS  \cite{bss}                & 56.154 ± 0.718 & 78.796 ± 0.131 & 87.170 ± 0.050 & 93.206 ± 0.044 \\ 
             ~              & Attention only (freeze FFN) \cite{3things} & 56.701 ± 0.521 & 77.872 ± 0.233 & 86.747 ± 0.256 & 92.414 ± 0.000 \\ 
             ~              & FFN only (freeze attention) \cite{3things} & 51.171 ± 0.799 & 75.418 ± 0.386 & 85.769 ± 0.273 & 92.671 ± 0.059 \\ 
             ~              & GTA                        & \textbf{59.271 ± 0.248} & \textbf{79.488 ± 0.202} & \textbf{87.651 ± 0.111} & 93.239 ± 0.097 \\
             \hline
             Aircraft       & Fine-tune (baseline)        & 52.115 ± 0.412 & 68.447 ± 0.647 & 76.848 ± 0.330 & 86.939 ± 0.076 \\
             ~              & $L_2$-SP \cite{l2sp}             & 51.645 ± 0.465 & 68.777 ± 0.666 & 76.978 ± 0.625 & \textbf{87.209 ± 0.121} \\ 
             ~              & BSS \cite{bss}              & 52.285 ± 0.291 & 68.677 ± 0.692 & 76.998 ± 0.330 & 87.129 ± 0.369 \\ 
             ~              & Attention only (freeze FFN) \cite{3things} & 50.735 ± 1.379 & 67.477 ± 0.505 & 76.098 ± 0.362 & 85.639 ± 0.522 \\ 
             ~              & FFN only (freeze attention) \cite{3things}& 51.195 ± 0.243 & 67.207 ± 0.390 & 75.198 ± 0.392 & 85.399 ± 0.809 \\ 
             ~              & GTA                        & \textbf{54.635 ± 0.572} & \textbf{70.027 ± 0.778} & \textbf{77.548 ± 0.632} & 86.989 ± 0.191 \\
             \hline
             Dogs           & Fine-tune (baseline)        & 59.775 ± 0.256 & 72.137 ± 0.220 & 78.131 ± 0.037 & 83.318 ± 0.007 \\ 
             ~              & $L_2$-SP \cite{l2sp}           & 63.893 ± 0.477 & 75.715 ± 0.603 & 81.453 ± 0.338 & 85.264 ± 0.186 \\ 
             ~              & BSS  \cite{bss}               & 59.817 ± 0.303 & 72.253 ± 0.087 & 78.155 ± 0.219 & 83.570 ± 0.251 \\ 
             ~              & Attention only (freeze FFN) \cite{3things} & 62.747 ± 0.455 & 74.577 ± 0.298 & 80.113 ± 0.114 & 84.938 ± 0.205 \\ 
             ~              & FFN only (freeze attention) \cite{3things} & 57.502 ± 0.299 & 70.194 ± 0.095 & 77.253 ± 0.125 & 83.182 ± 0.273 \\ 
             ~              & GTA                        & \textbf{69.196 ± 0.222} & \textbf{78.054 ± 0.194} & \textbf{81.803 ± 0.036} & \textbf{85.633 ± 0.192} \\
             \hline
             Pet            & Fine-tune (baseline)        & 77.342 ± 0.382 & 86.418 ± 0.433 & 90.206 ± 0.096 & 93.123 ± 0.201 \\ 
             ~              & $L_2$-SP \cite{l2sp}          & 81.185 ± 0.500 & 88.871 ± 0.220 & 92.169 ± 0.299 & \textbf{94.276 ± 0.439} \\ 
             ~              & BSS   \cite{bss}            & 77.478 ± 0.488 & 86.572 ± 0.450 & 90.597 ± 0.206 & 93.286 ± 0.417 \\ 
             ~              & Attention only (freeze FFN) \cite{3things} & 81.030 ± 0.666 & 88.698 ± 0.259 & 91.832 ± 0.306 & 93.786 ± 0.166 \\ 
             ~              & FFN only (freeze attention) \cite{3things} & 74.825 ± 0.886 & 84.755 ± 0.129 & 89.697 ± 0.382 & 92.723 ± 0.142 \\ 
             ~              & GTA                        & \textbf{83.856 ± 0.063} & \textbf{89.906 ± 0.197} & \textbf{92.478 ± 0.245} & 94.022 ± 0.246 \\
             \bottomrule
     \end{tabular}
     \vspace{-4pt}
     \caption{\textbf{Comparison of transfer learning methods.} The baseline refers to the na\"{\i}vely fine-tuned model. 
 ``Attention only'' and ``FFN only'' represent training of only attention layers and feed-forward network (FFN), respectively. GTA shows higher accuracy across all datasets and all sampling rates, with particularly significant improvements when the training data is limited. The best results are bold-faced.}
     \vspace{-5pt}
     \label{tbl:main_table}
\end{center}
\end{table*}

\section{Experimental Results}

In this section, we evaluate the effectiveness of our method on several fine-grained datasets, which serve as standard benchmarks for assessing TL performance. Our experiments highlight the importance of applying regularization to the attention logits of the \texttt{[cls]} token. We also present segmentation results that show how the attention logits of the target model focus on objects that are relevant to the target task, rather than simply duplicating those of the source model. Furthermore, we evaluate the synergies between our method and the recently developed augmentation technique TransMix~\cite{transmix}, which exploits the attention outputs in ViT. Finally, we conduct an ablation study to investigate the impact of key factors on the performance of our proposed method.

\label{sec:experiments}

\paragraph{Datasets.}
We employ five widely used fine-grained datasets: CUB-200-2011 (CUB)~\cite{cub200}, Stanford Cars (Cars)~\cite{stanfordcars}, FGVC-Aircraft (Aircraft)~\cite{aircraft}, Stanford Dogs (Dogs)~\cite{stanforddogs}, and Oxford-IIIT Pet (Pet)~\cite{oxfordiiitpets}, which contain birds, cars, airplanes, dogs, and pets, respectively. Table~\ref{tbl:dataset} shows the data statistics for the datasets. We conduct experiments with four different configurations based on the amount of training data following~\cite{bss, co-tuning}. Each configuration consists of a varying percentage of randomly selected training samples for each category: 15\%, 30\%, 50\%, and 100\%. These datasets for fine-grained classification have been extensively studied in TL~\cite{bss,co-tuning,delta,l2sp}.

\paragraph{Training configurations.}
We follow DINO fine-tuning configurations~\cite{dino} and apply them to all methods, including the baseline (i.e., na\"{\i}ve fine-tuning). All methods are trained using AdamW optimizer with a momentum of 0.9 during 3k iterations, and the learning rate is decreased by cosine annealing scheduler~\cite{cosine}. We set the batch size, weight decay, and initial learning rate to 768, 0.05, and 0.0001, respectively. The input images are resized to 224$\times$224. RandAugment~\cite{randaugment} is employed for augmentation. However, we do not use random erasing~\cite{randomerasing} since self-attention layers strongly focus on the areas randomly erased, which can lead to inaccurate attention guidance. All experiments are conducted with the ViT-small architecture. All weights are initialized with the ImageNet-1k pre-trained checkpoint of iBOT. We repeat each experiment three times with different random seeds to report performance variations.

\subsection{Transfer Learning Performance}
Firstly, we compare our method with previous TL methods (see Table~\ref{tbl:main_table}) to verify their compatibility with ViT. Also, we evaluate the effectiveness of GTA in leveraging object-centric representations. To make the comparison as fair as possible, we mostly use the hyperparameter settings reported in each paper, but a regularization coefficient $\lambda$ is tested with three values based on the default values of each TL method. 
Specifically, we train models with 0.1$\times \alpha$, $\alpha$, and 10$\times \alpha$ when $\alpha$ is the default value. We report the best performance among the results obtained using three different $\lambda$ values.

At the lowest sampling rate setting (i.e. 15\%), GTA can significantly enhance performance compared to the baseline for all datasets. Specifically, each dataset shows an improvement of at least 2.52\% and up to 10.15\%. When the training data is insufficient, ViT tends to attend more to the background rather than the foreground objects, making it challenging to classify images with different backgrounds in the test dataset. However, GTA addresses this issue by explicitly regularizing the attention on foreground objects. As the amount of training data increases, the degree of improvement decreases. For example, with the CUB dataset, the gaps between GTA and baseline are reduced to 15\%: 10.149, 30\%: 5.719, 50\%: 2.900, and 100\%: 1.099.

We also compare GTA with commonly used TL methods such as $L_2$-SP~\cite{l2sp}, BSS~\cite{bss}, and ViT-specific methods~\cite{3things}. Our results demonstrate that GTA consistently outperforms the comparison methods at almost all sampling rates, especially in cases where the training dataset is relatively small. Across all target datasets, the gap between GTA and the best-performing previous TL methods ranges from 2.35\% to 8.89\% at the 15\% setting. While this result can be consistently observed at the 30\% and 50\% settings, the performance gap between GTA and other methods decreases, eventually becoming comparable at the 100\% setting. For instance, The $L_2$-SP shows comparable results with GTA at the 100\% configuration for Cars, Aircraft, and Pet datasets. 

The $L_2$-SP is the most explicit and simplest method to take advantage of a well-trained source model. However, we observe that combining $L_2$-SP with ViT does not lead to a consistent performance improvement.
The BSS method has the advantage of excluding negative features from the pre-trained model, but it lacks regularization terms to leverage transferred knowledge, making it prone to overfitting to the target task, similar to the baseline. According to~\cite{3things}, training only attention layers yields better performance than end-to-end fine-tuning. While it is also observed in our experiments, the method shows lower performance than GTA. Similarly, the FFN-only method, which freezes the attention layers from the pre-trained model, shows poor performance since the frozen attention cannot be adapted to the target task.

\subsection{The Importance of Attention Logits} 
\label{subsec:guide_importance}
\begin{table}
\small
\centering
\newcolumntype{L}[1]{>{\raggedright\let\newline\\\arraybackslash\hspace{0pt}}m{#1}}
\newcolumntype{C}[1]{>{\centering\let\newline\\\arraybackslash\hspace{0pt}}m{#1}}
\newcolumntype{R}[1]{>{\raggedleft\let\newline\\\arraybackslash\hspace{0pt}}m{#1}}
\begin{tabular}{L{1.3cm}L{2.9cm}R{1.2cm}R{1.2cm}} 
\toprule
                           &                          & \multicolumn{2}{c}{\textbf{Sampling Rates}}  \\
\textbf{\textbf{Dataset}} & \textbf{\textbf{Method}} & \textbf{15\%}   & \textbf{100\%}              \\ 
\hline
CUB                       & baseline                 & 41.376          & 84.444                      \\
                          & block output guide       & 46.859          & 85.077                      \\
                          & MSA output guide         & 46.519          & 84.904                      \\
                          & Attention logits (GTA)                      & \textbf{51.525}          & \textbf{85.543}                      \\ 
\hline
Cars                      & baseline                 & 56.100          & 93.065                      \\
                          & block output guide       & 58.960          & 93.098                      \\
                          & MSA output guide         & 59.039          & 93.023                      \\
                          & Attention logits (GTA)             & \textbf{59.271} & \textbf{93.239}             \\ 
\hline
Aircraft                  & baseline                 & 52.115          & 86.939                      \\
                          & block output guide       & 54.485          & 86.999                      \\
                          & MSA output guide         & 54.225          & \textbf{87.039}                      \\
                          & Attention logits (GTA)             & \textbf{54.635} & 86.989             \\ 
\hline
Dogs                      & baseline                 & 59.775          & 83.318                      \\
                          & block output guide       & 65.299          & 84.755                      \\
                          & MSA output guide         & 65.078          & 84.740                      \\
                          & Attention logits (GTA)             & \textbf{69.196} & \textbf{85.633}             \\ 
\hline
Pet                       & baseline                 & 77.342          & 93.123                     \\
                          & block output guide       & 82.875          & 93.913                      \\
                          & MSA output guide         & 82.666          & 93.877                      \\
                          & Attention logits (GTA)             & \textbf{83.856} & \textbf{94.022}             \\
\bottomrule
\end{tabular}
\vspace{-4pt}
\caption{\textbf{Effectiveness of different features for guidance.} The block output and MSA output guide indicate the guidance between source and target model with the transformer block output and the MSA layer output, respectively. Our proposed method, GTA, provide guidance to target model using attention logits. The proposed method shows higher accuracy across all dataset and sample rates. Best results are bold-faced.}
\label{tbl:_guide}
\end{table}

Table~\ref{tbl:_guide} shows the effectiveness of guiding attention logits, particularly when contrasted with the utilization of two other outputs, the transformer block output $\mathbf{z}^{\prime}$ and MSA output $\operatorname{MSA}(\mathbf{z})$ of the ViT architecture. To ensure a comprehensive evaluation, we apply $L_2$ regularization to these alternative outputs following Equation~\ref{eqn:guide}. Our experiments confirm that GTA outperforms the regularization of other outputs across different sampling rates and datasets. For example, the performance gaps are in the range of 0.15\% and 5.01\% at the 15\% sampling rate.
This tendency has been similarly observed at 30\% and 50\% settings. These results reveal the crucial importance of selecting attention logits for the guiding mechanism, implying that alternatives may causally lead to negative transfer. By leveraging attention logits for guidance, our approach mitigates the risk of such undesirable consequences. 
It is important to note that while the guidance provided by attention logits does not explicitly regularize the trained \emph{features} (i.e., the MSA output or block output), it corresponds to an effective inductive bias rooted in well-trained kernels.
Such a bias strategically directs the spatial attention towards foreground areas, thereby increasing the accuracy of the classification task.

\begin{table}[h]
\centering
\newcolumntype{L}[1]{>{\raggedright\let\newline\\\arraybackslash\hspace{0pt}}m{#1}}
\newcolumntype{C}[1]{>{\centering\let\newline\\\arraybackslash\hspace{0pt}}m{#1}}
\newcolumntype{R}[1]{>{\raggedleft\let\newline\\\arraybackslash\hspace{0pt}}m{#1}}
\begin{tabular}{L{4cm}R{3.5cm}} 
\toprule
\textbf{Method}          & \textbf{Jaccard index}  \\ 
\hline
baseline                 & 0.367                    \\
pre-trained (SSL)        & 0.386                    \\
GTA                      & \textbf{0.399}           \\
\bottomrule
\end{tabular}
\vspace{-4pt}
\caption{\textbf{Quantitative evaluation of attention map guidance on segmentation task.} Baseline refers to simple fine-tuning, pre-trained denotes SSL models not yet train for the target task. The proposed GTA outperformed the others in terms of Jaccard index on PASCAL-VOC12 validation set. Best results are bold-faced.}
\vspace{0pt}
\label{tbl:seg}
\end{table}

\subsection{Segmentation Performance} 

\begin{figure}[t!]
    \centering
    \includegraphics[width=6cm]{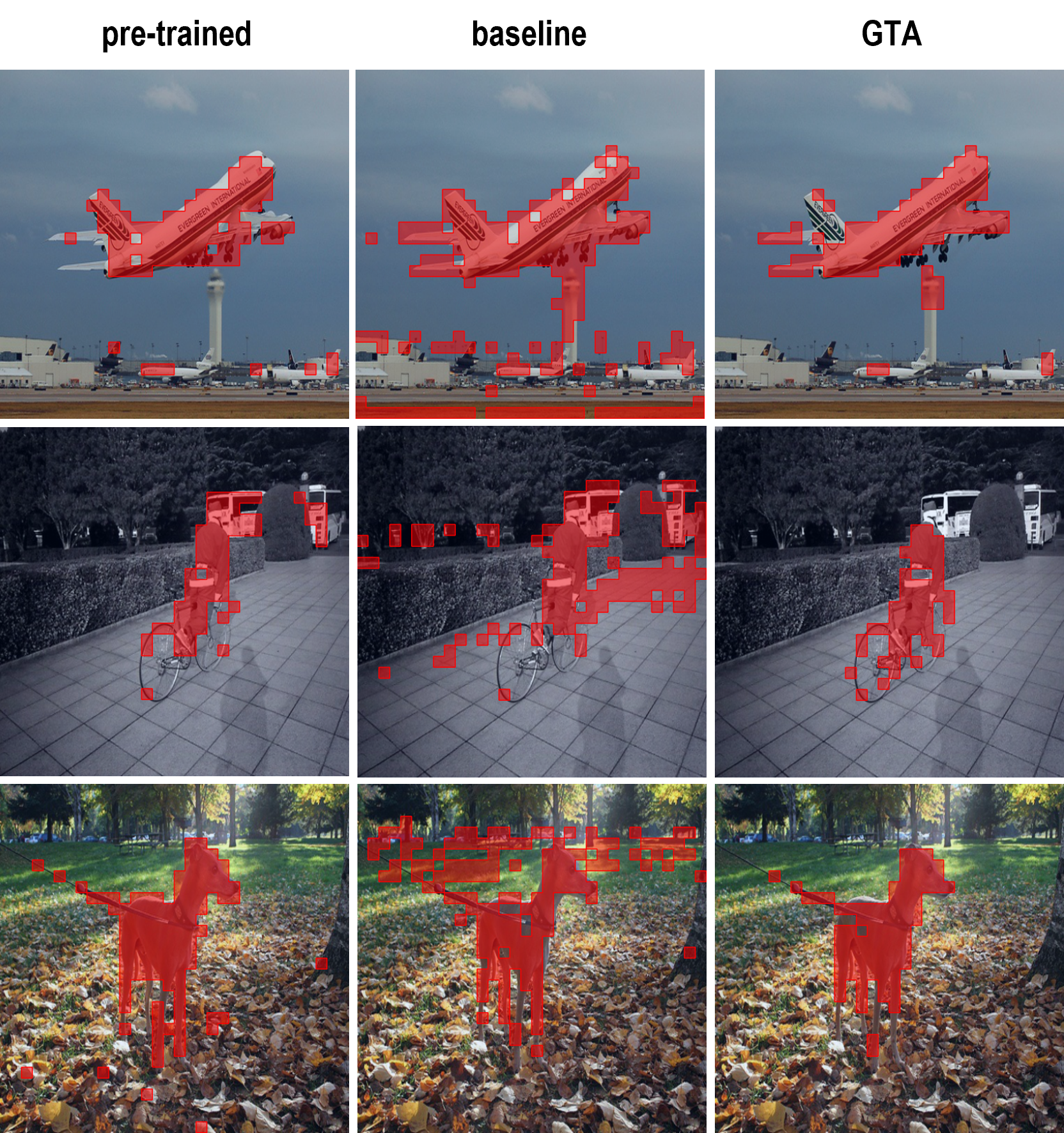}
    \vspace{+3pt}
   \caption{\textbf{Comparison of segmentation results on PASCAL-VOC12.} Pre-trained refers to the segmentation results obtained by the attention logits of the upstream. The baseline represents the results obtained by fine-tuning the pre-trained model to target task. GTA denotes the results obtained by utilizing the GTA during fine-tuning. GTA shows optimized performance compared to the other results.}
  \vspace{-10pt}
\label{fig_seg}
\end{figure}

In this experiment, we compare the segmentation results obtained from the GTA model with those of the SSL source model and fine-tuned model by evaluating segmentation performance on the PASCAL-VOC12 validation set using the Jaccard index~\cite{voc}, following~\cite{dino, ibot, intriguing}. The visualization results show that the segmentation results from GTA are more accurate in focusing on the foreground object, as shown in Figure~\ref{fig_seg}. Quantitatively, the GTA model also shows a higher Jaccard index compared to others (see Table~\ref{tbl:seg}). The fine-tuned model focuses on specific parts of the foreground but also attends to a significant amount of irrelevant background information. The SSL model performs well, but also places attention on unimportant areas that are not relevant to the target class. While the segmentation results generated by GTA do not perfectly replicate those of the SSL model, it effectively focuses on informative areas of the target object while ensuring that the model is optimized for the current target task.

\subsection{Boosting Effect of Attention Guidance} 

\begin{table}
\small
\centering
\newcolumntype{L}[1]{>{\raggedright\let\newline\\\arraybackslash\hspace{0pt}}m{#1}}
\newcolumntype{C}[1]{>{\centering\let\newline\\\arraybackslash\hspace{0pt}}m{#1}}
\newcolumntype{R}[1]{>{\raggedleft\let\newline\\\arraybackslash\hspace{0pt}}m{#1}}
\begin{tabular}{L{1.3cm}L{2.9cm}R{1.2cm}R{1.2cm}} 
\toprule
                 &                     & \multicolumn{2}{c}{\textbf{Sampling Rates}}  \\ 
\textbf{Dataset} & \textbf{Method}     & \textbf{15\%} & \textbf{100\%}                \\ 
\hline
CUB              & baseline            & 41.376        & 84.444                        \\
                 & baseline + TransMix & 42.032        & 84.703                        \\
                 & GTA                & 51.525        & 85.543                        \\
                 & GTA + TransMix     & \textbf{54.361}        & \textbf{85.755}                        \\
\hline
Cars             & baseline            & 56.100        & 93.065                        \\
                 & baseline + TransMix & 56.117        & 93.139                        \\
                 & GTA                & 59.271        & \textbf{93.239}                        \\
                 & GTA + TransMix     & \textbf{59.943}        & 93.218                        \\
\hline
Aircraft         & baseline            & 52.115        & 86.939                        \\
                 & baseline + TransMix & 52.455        & 86.819                        \\
                 & GTA                & 54.635        & 86.989                        \\
                 & GTA + TransMix     & \textbf{55.166}        & \textbf{87.369}                        \\
\hline
Dogs             & baseline            & 59.775        & 83.318                        \\
                 & baseline + TransMix & 60.229        & 83.551                        \\
                 & GTA                & 69.196        & 85.633                        \\
                 & GTA + TransMix     & \textbf{70.004}        & \textbf{85.793}                        \\
\hline
Pet              & baseline            & 77.342        & 93.123                        \\
                 & baseline + TransMix & 77.396        & 93.268                        \\
                 & GTA                & 83.856        & 94.022                        \\
                 & GTA + TransMix     & \textbf{84.937}        & \textbf{94.067}                        \\
\bottomrule
\end{tabular}
\vspace{-4pt}
\caption{ \textbf{Quantitative evaluation of the boosting effect.} Baseline refers to the fine-tuned model without TransMix or GTA. +TransMix denote add TransMix augmentation on tranining. The combination of GTA and TransMix outperformed both the baseline and GTA alone. Best results are bold-faced.}
\vspace{-3pt}
\label{tbl:boost}
\end{table}

As demonstrated in our previous experiment, we show that GTA improves the localization quality of the self-attention logits on the target object. To capitalize on this advantage, we investigate whether a boosting effect can be achieved by combining GTA with TransMix~\cite{transmix}. TransMix mixes images in a similar manner to CutMix~\cite{cutmix}, but without using the size ratio of the cropped box as a new label. Instead, a new label is calculated based on the self-attention ratio between the mixed images. The effectiveness of TransMix relies on the ability of the target model to generate proper attention that is accurately focused on the foreground object. However, the authors observed that an attention map that accurately localizes objects does not helpful to improve the performance of TransMix through the experiments using DINO as a parameter-frozen external model. The parameter-frozen external model has a limitation in that it can only generate mixing labels in a static manner, regardless of training progress. In contrast, our proposed method allows for dynamic mixing labels while incorporating improved attention from an external model since the parameter-frozen external model guides only the attention logit of the target model. 

According to Table~\ref{tbl:boost}, TransMix shows better performance when it is combined with GTA rather than when it is used with the baseline. The performance gap between baseline and baseline+TransMix and that between GTA and GTA+TransMix is significantly increased when the sampling rate is small. When training with a small dataset, the background attention issue, as visualized in Figure~\ref{fig1_intro}, can hinder TransMix from generating the appropriate labels. However, as the amount of training data increases, the effect of attention improvement by GTA decreases, and consequently the boosting effect is also reduced. Since the combination of TransMix and GTA shows better results than GTA alone, it demonstrates that GTA can be combined with other regularization methods to further improve the results.

\subsection{Ablation Study} 

The performance of GTA can be influenced by two main factors: the selection of the pre-trained weight used as the source model and the appropriate regularization coefficient $\lambda$. In this section, we analyze these factors in detail. 

\paragraph{Selection of guidance model.} 
GTA is the method that guides the training of the target model using the source model. Therefore, the choice of which weights to use as the source model can affect the performance of GTA. In this experiment, we compare the performance of using SSL models (DINO and iBOT) and the commonly used SL model (ImageNet-1k) as the source model. Our results show that GTA consistently improves accuracy across all datasets, whether applied to SL or SSL (see Table~\ref{tbl:weight} for the comparison with the SL model and Appendix A for DINO experiments). This suggests that GTA is not dependent on specific SSL weights, but rather can be applied to a variety of pre-trained models. However, there are performance differences depending on which weights are used. When using SL weights, we observe better performance on CUB, Dogs, and Pet datasets, whereas when using SSL weights, we observe better results on Cars and Aircraft compared to SL. These differences can be attributed to domain discrepancies between upstream and downstream data~\cite{imagenetbetter}. Since the SL model is trained on ImageNet for classification, CUB, Dogs, and Pet are semantically close to the upstream domain, while Car and Aircraft are not, resulting in lower baseline performance. In contrast, the SSL models show better generalization performance, leading to better results on Cars and Aircraft despite the fact that SSL is also trained on ImageNet. 
 
\begin{table}[]
\small
\centering
\newcolumntype{L}[1]{>{\raggedright\let\newline\\\arraybackslash\hspace{0pt}}m{#1}}
\newcolumntype{C}[1]{>{\centering\let\newline\\\arraybackslash\hspace{0pt}}m{#1}}
\newcolumntype{R}[1]{>{\raggedleft\let\newline\\\arraybackslash\hspace{0pt}}m{#1}}
\begin{tabular}{L{1.3cm}L{2.9cm}R{1.2cm}R{1.2cm}}
\toprule
                 &                    & \multicolumn{2}{c}{               \textbf{Sampling Rates}}    \\
\textbf{Dataset} & \textbf{Method}    & \textbf{15\%}                & \textbf{100\%}                 \\ \hline
CUB     & baseline (SL)  & 51.519 & 85.548  \\
                 & GTA (SL)       & \textbf{62.047} & \textbf{85.663}  \\
                 & baseline (SSL) & 41.376 & 84.444  \\
                 & GTA (SSL)      & 51.525 & 85.543  \\ \hline
Cars    & baseline (SL)  & 45.894 & 91.382  \\
                 & GTA (SL)       & 47.822 & 90.930  \\
                 & baseline (SSL) & 56.100 & 93.065  \\
                 & GTA (SSL)      & \textbf{59.271} & \textbf{93.239}  \\ \hline
Aircraft         & baseline (SL)  & 48.355 & 82.638  \\
                 & GTA (SL)       & 49.635 & 82.558  \\
                 & baseline (SSL) & 52.115 & 86.939  \\
                 & GTA (SSL)      & \textbf{54.635} & \textbf{86.989}  \\ \hline
Dogs    & baseline (SL)  & 74.872 & 87.945  \\
                 & GTA (SL)       & \textbf{88.897} & \textbf{91.682}  \\
                 & baseline (SSL) & 59.775 & 83.318  \\
                 & GTA (SSL)      & 69.196 & 85.633  \\ \hline
Pet & baseline (SL)  & 81.466 & 93.123  \\
                 & GTA (SL)       & \textbf{91.524} & \textbf{94.967}  \\
                 & baseline (SSL) & 77.342 & 93.123  \\
                 & GTA (SSL)      & 83.856 & 94.022 \\
\bottomrule
\end{tabular}
\vspace{+5pt}
\caption{ \textbf{Comparison of GTA performance using different source model weights.} GTA consistently improved accuracy on all datasets using both SSL and SL weights as the source model. Best results are bold-faced.}
\vspace{-10pt}
\label{tbl:weight}
\end{table}

\paragraph{Influence of $\lambda$.} 

\begin{figure}
    \centering
    \includegraphics[width=8.2cm]{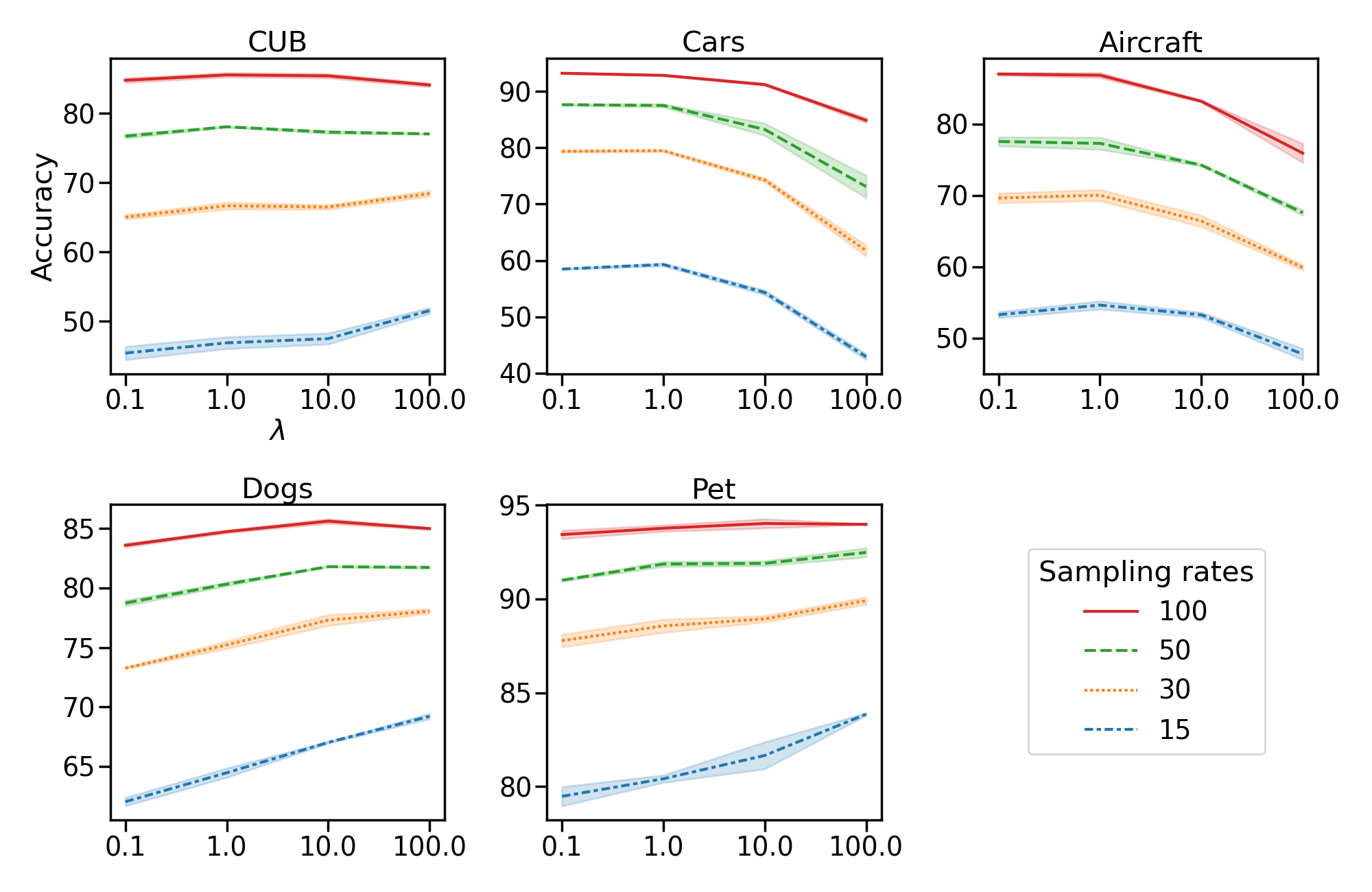}
   \caption{\textbf{The effect of different values of $\lambda$ on GTA.} The optimal lambda value varies depending on the characteristics and amount of the target data.}
   \vspace{-10pt}
\label{fig_lambda}
\end{figure}

We test four different $\lambda$ values (0.1, 1.0, 10.0, 100.0) to find an optimal value for each dataset (see Figure~\ref{fig_lambda}). Our findings reveal that the optimal $\lambda$ varies depending on the size and characteristics of the dataset. Similar to the weight experiments above, we observe that the results of $\lambda$ are also strongly influenced by the characteristics of the data domain. Specifically, datasets such as CUB, Dogs, and Pet that belong to the domain close to the upstream data (called the near-domain) show good performance with high $\lambda$ values. In contrast, datasets such as Cars and Aircraft, belonging to the domain semantically far from the upstream data (called the out-domain), show better results with low $\lambda$ values. The difference could be attributed to the quality of the self-attention logits used for guidance. In the case of near-domain, even with high $\lambda$, the target task can be learned well with minimal changes in the self-attention logits. However, in the out-domain, a considerable change in the self-attention logits is required to learn the target task. Therefore, as the target data are far from the upstream data domain, smaller $\lambda$ values should be used, but too small $\lambda$ values could lead to shortcut learning similar to the baseline fine-tuning. As a result, our experiments show that for out-domain datasets, the optimal value of $\lambda$ is consistently 1.0 regardless of the amount of training data. In contrast, a higher value of $\lambda$ yields better accuracy as the amount of data decreases for near-domain datasets. At the 15\% setting, $\lambda=100.0$ is preferred, but for higher sampling ratios, $\lambda=10.0$ is found to be the optimal value. Hence, when applying GTA, it is necessary to set a parameter $\lambda$ based on the characteristics and the amount of target data.

\section{Conclusion}
In this paper, we propose a novel transfer learning method called GTA, which effectively utilizes pre-trained knowledge to improve TL performance, specifically for the ViT architecture. By applying explicit $L_2$ regularization between the attention logits of the target and source models, GTA can achieve significant performance improvements across various fine-grained datasets and sampling rates. Through extensive experiments, we show that imposing regularization on the attention logits in ViT is essential, and that GTA outperforms other comparison methods especially when the number of target training data is small. These results demonstrate that GTA is a simple and effective approach to improve the TL performance of ViT.

{\small
\bibliographystyle{ieee_fullname}
\bibliography{main.bbl}
}

\clearpage
\begin{appendix}

\renewcommand{\thesection}{\Alph{section}}

\renewcommand\thefigure{\arabic{figure}}
\renewcommand\thetable{\arabic{table}}

\setcounter{figure}{0}
\setcounter{table}{0}

\section*{Appendix}

\section{Effect of SSL guidance models}
We conducted additional evaluations on the effect of different SSL weights for guidance. We selected DINO as a comparative benchmark, which is widely used and has excellent attention localization performance~\cite{dino}. We compared the performance of GTA with both DINO and iBOT. Both weights showed improved performance, but iBOT exhibited even greater improvement. This is because iBOT has superior localization performance than DINO~\cite{ibot}, leading to more accurate attention guidance (see Table ~\ref{tbl:sup_weight}).

\begin{table}[h!]
\vspace{10pt}
\small
\centering
\newcolumntype{L}[1]{>{\raggedright\let\newline\\\arraybackslash\hspace{0pt}}m{#1}}
\newcolumntype{C}[1]{>{\centering\let\newline\\\arraybackslash\hspace{0pt}}m{#1}}
\newcolumntype{R}[1]{>{\raggedleft\let\newline\\\arraybackslash\hspace{0pt}}m{#1}}
\begin{tabular}{L{1.3cm}L{2.9cm}R{1.2cm}R{1.2cm}}
\toprule
                 &                    & \multicolumn{2}{c}{               \textbf{Sampling Rates}}    \\
\textbf{Dataset} & \textbf{Method}    & \textbf{15\%}                & \textbf{100\%}                 \\ \hline
CUB              & baseline (DINO) & 38.310 & 83.512  \\
                 & GTA (DINO)      & 48.320 & 84.711  \\
                 & baseline (iBOT) & 41.376 & 84.444  \\
                 & GTA (iBOT)      & \textbf{51.525} & \textbf{85.543}  \\ \hline
Cars             & baseline (DINO) & 52.688 & 92.741  \\
                 & GTA (DINO)      & 56.150 & 92.886  \\
                 & baseline (iBOT) & 56.100 & 93.065  \\
                 & GTA (iBOT)      & \textbf{59.271} & \textbf{93.239}  \\ \hline
Aircraft         & baseline (DINO) & 51.055 & 85.649  \\
                 & GTA (DINO)      & 53.335 & 86.269  \\
                 & baseline (iBOT) & 52.115 & 86.939  \\
                 & GTA (iBOT)      & \textbf{54.635} & \textbf{86.989}  \\ \hline
Dogs             & baseline (DINO) & 57.207 & 82.778  \\
                 & GTA (DINO)      & 66.099 & 84.705  \\
                 & baseline (iBOT) & 59.775 & 83.318  \\
                 & GTA (iBOT)      & \textbf{69.196} & \textbf{85.633}  \\ \hline
Pet              & baseline (DINO) & 75.034 & 92.596  \\
                 & GTA (DINO)      & 80.113 & 94.022 \\
                 & baseline (iBOT) & 77.342 & 93.123  \\
                 & GTA (iBOT)      & \textbf{83.856} & \textbf{94.022} \\

\bottomrule
\end{tabular}
\vspace{+3pt}
\caption{ \textbf{Comparison of GTA performance using different SSL weights.} GTA consistently improved accuracy on all datasets using both DINO and iBOT weights as the source model. Best results are bold-faced.}
\vspace{5pt}
\label{tbl:sup_weight}
\end{table}

\section{Comparison of self-attention maps}
\begin{figure*}[h!]
    \centering
    \includegraphics[width=15.0cm]{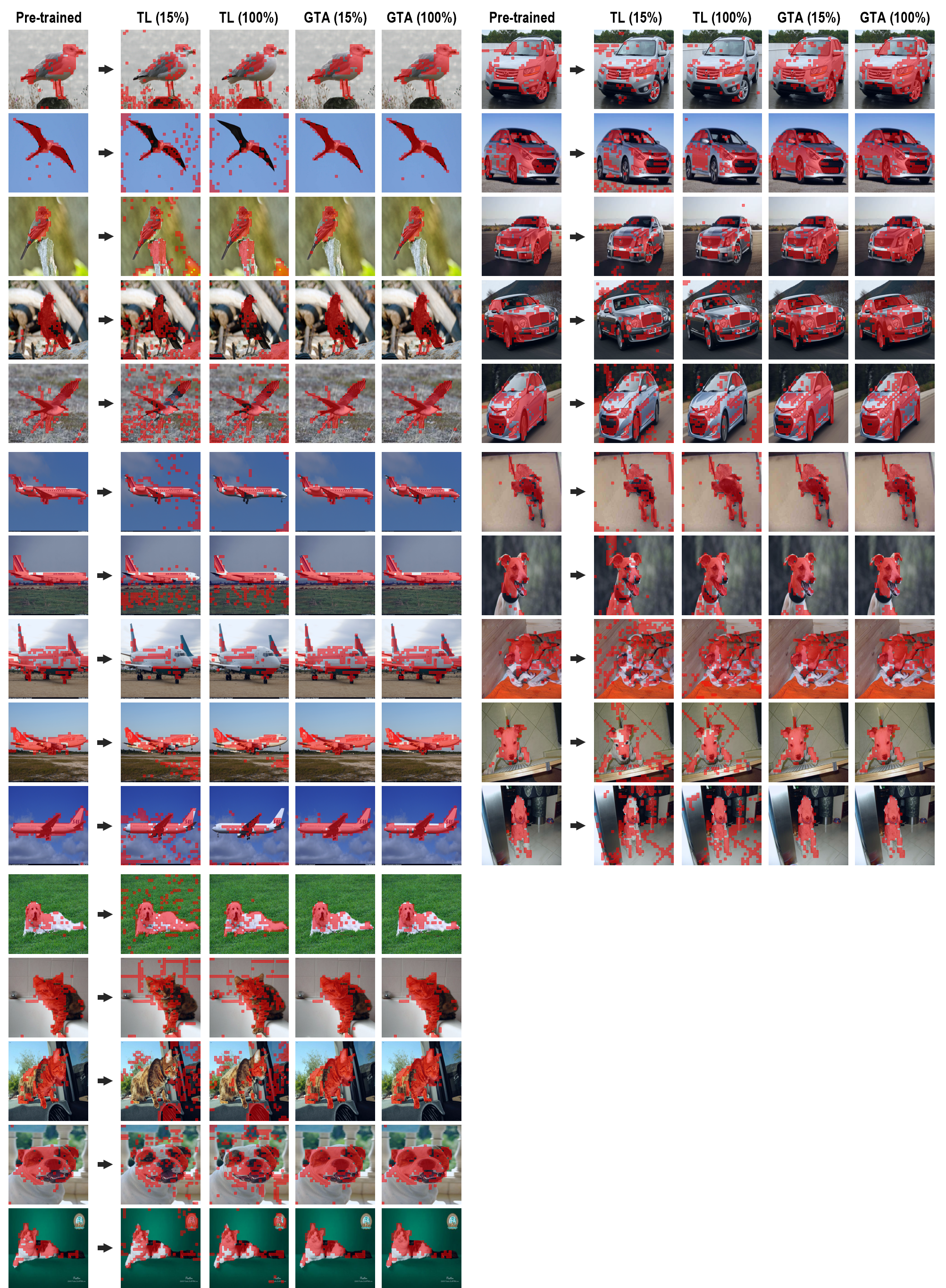}
   \caption{\textbf{Comparison of self-attention maps from pre-trained, na\"{\i}vely fine-tuned, and GTA-traind models across multiple datasets.} We consider CUB, Cars, Aircraft, Dogs, and Pets datasets. The self-attention maps of the multiple heads are aggregated with maximum values, and visualized in red color. Each column shows the attention maps from the models that are pre-trained using SSL, fine-tuned, and fine-tuned with GTA on 15\% and 100\% of training data, respectively. GTA shows that it is capable of fully leveraging object-centric representations learned by the SSL model.}
\label{sup_fig}
\end{figure*}

In this section, we show additional visual comparisons of the self-attention maps obtained from pre-trained, fine-tuned, and GTA-trained models on multiple datasets (see Figure \ref{sup_fig}) \cite{stanforddogs, stanfordcars, aircraft, oxfordiiitpets, cub200}. The self-attention maps allow us to understand where the model attends to different parts of the input image.

For each dataset, we randomly select a sample image and visualize the self-attention maps. We observe that the self-attention maps of the fine-tuned model are much scattered over non-meaningful areas, in contrast to the pre-trained model which demonstrates focused attention on important regions. Such behavior could lead to the loss of well-trained spatial information, eventually resulting in lower performance. However, by introducing GTA, we show that it is possible to avoid this issue by explicitly regularizing the attention logits between target and source models. We present a visual comparison of the self-attention maps from these models to illustrate the effectiveness of the proposed method in guiding attention towards important regions during training. The visualization results demonstrate that GTA-trained models outperform fine-tuned models on multiple datasets.

\end{appendix}

\clearpage

\end{document}